# A Framework for Comparing Uncertain Inference Systems to Probability

Ben P. Wise & Max Henrion
Robotics Institute and
Dept. of Engineering and Public Policy
Carnegie-Mellon University
Pittsburgh, Pa 15213

## Introduction

Uncertainty is a salient feature of most domains of expert knowledge and of the problems to which it is applied. Developers of rule-based expert systems have employed several different techniques to represent uncertainty and make inferences under uncertainty. We shall refer to these techniques generically as uncertain inference systems (UIS's).

Probability is by far the most extensively developed formalism for representing uncertainty, but the majority of researchers in rule-based expert systems have not found it very appealing hitherto. Instead they have developed more ad-hoc techniques, notably Certainty Factors used in Mycin (Shortliffe, 1976), and the related approach used in Prospector (Duda et al, 1976). Other methods in which there is current interest are Fuzzy Set Theory (Zadeh, 1984, Kaufmann, 1975), the belief functions of Dempster-Shafer theory, Cohen's theory of endorsements, and Doyle's theory of reasoned assumptions. There is increasing debate about the relative merits and flaws of these approaches (e.g. see (Cohen et al, 1984, Spiegelhalter, 1985)), and the question arises of how to decide which is most appropriate for a given application.

A major difficulty in comparing them is that they make quite different fundamental assumptions about the nature of uncertainty. Where the notion of subjective probability in Bayesian Decision Theory refers to the degree of belief in a hypothesis, Fuzzy Set theory addresses vagueness or linguistic imprecision, and Dempster-Shafer theory addresses the degree of evidential support or disconfirmation. They also have varying types of theoretical status. Probability theory is based on explicit axioms of rational choice behavior; whereas other UISs, such as Mycin's Certainty Factors, have only heuristic justification, seeming to be useful even though they have do not have a precise, operationally defined meaning.

The second set of issues in comparing UISs are pragmatic. How easy is it to implement? What kinds of knowledge structures does it imply? How easy is to elicit the expert judgments required in this form? How easy is to understand the conclusions of uncertain inferences? And not least, how much computational effort does it require? In keeping with the pragmatism dominant in research in AI and expert systems, these issues, particularly the last, seem to have been the prime determinants so far in choice of UIS.

Researchers in AI have given more attention to qualitative issues of symbolic reasoning, than issues of quantification of uncertainty. Indeed there is a common perception that results are insensitive to the method of quantification employed, although we have been unable to find experimental tests of this belief. Thus a third type of question in selecting among UIS is about their performance: Does it make a significant difference in terms of the conclusions of uncertain inference which approach you use? If not, then we may continue to use the simpler methods such as in Mycin, without having to worry about the complexities of some of the other UISs. But if there are situations in which commonly used UISs produce clearly incorrect or counter-intuitive results, then at least we should be aware of these so we can avoid them. If these situations are common then perhaps we should change our choice of UIS, or identify a need to develop better methods which combine greater theoretical soundness with pragmatic ease-of-use. Until more thorough tests of the performance of current UIS's are carried out, we cannot know how much of a problem there is, and so it is hard to tell the degree of urgency. The main goal of this paper is to present a framework for making such experimental comparisons, some analysis of the behavior of selected UIS's for rule-components and some preliminary results with an example rule-set.

## Formalisms for representing uncertainty

In this section we will provide an extremely brief introduction to the most popular formalisms for representing uncertainty, including probability, as used in Bayesian Decision Theory (BDT), Mycin certainty factors (Myc), Fuzzy Set Theory (FST), and Dempster-Shafer Theory (DST).

### Probability

Probability is certainly the oldest and most widely used formalism for representing uncertainty, at least outside Artificial Intelligence research. The probability of a proposition, according to Bayesian Decision Theory (BDT), is a measure of a person's degree of belief in it, given the information currently known to that person. It is defined operationally in terms of the person's willingness to take bets based on the truth of the proposition. Thus the notion of probability derives from a set of simple axioms about decision-making (Savage, 1954). The force of these axioms, and hence of the laws of probability derived from them, arises from the fact that a people who violate them and act on "incoherent" probabilities (which, for example, do not satisfy Bayes' rule) are liable to demonstrable loss. Most famously, an opponent could always design a "Dutch book", that is a combination of bets that they would be willing to accept, according to their professed beliefs, but which in sum would result in a guaranteed loss.

It has also been shown that for any reasonable scoring rule (which rewards a decision maker based on the truth of uncertain propositions), any scalar measure of uncertainty is either worse than probability (produces a lower score) or is equivalent to it (de Finetti, 1974).

It is important to note that these proofs are prescriptive, not descriptive. They do not claim that people actually behave according to the tenets of BDT or use subjective probabilities, but only that they could not do better than to follow it. Indeed there is a considerable psychological literature showing that human judgment is liable to predictable biases and inconsistencies arising from the cognitive heuristics we use (Kahneman, Slovic & Tversky, 1982).

A set of m *propositions*, $\{A_1, A_2, \ldots A_m\}$, each of which may be true or false, gives rise to $2^m$ different possible elementary *events*, each being a particular combination of propositions, e.g. $(A_1 \& \neg A_2 \& \ldots A_m)$. A full joint probability distribution over these propositions specifies a probability for each event, and so requires specification of $2^m$-1 parameters. (The extra parameter is taken up by the constraint that the probabilities sum to 1.)

A rule such as, $A_1 \& \neg A_2 \rightarrow A_3$ with probability p, has a natural interpretation as a conditional probability: $p(A_3 | A_1, \neg A_2) = p$ Such conditional probabilities and marginal probabilities, $p(A_i)$, representing prior opinions on events, impose a set of linear constraints on the joint probability distribution over the propositions. The exponentially large number of degrees of freedom to be specified and computational effort in updating distributions when new evidence becomes available make it quite understandable that expert system research has shied away from this full probabilistic representation.

### Mycin certainty factors and Fuzzy Set theory

Probably the earliest and most widely-used method for inference under uncertainty in expert systems are the Certainty Factors originally developed for Mycin (Shortliffe, 1976) and increasingly used in other systems. These were introduced as a computationally simpler alternative to subjective probabilities. They represent degree of belief in a proposition by a number between -1 and 1: 1 represents "certain truth", 0 means "no evidence", and -1 means "certain falsity".

Fuzzy set theory (FST) (Kaufmann, 1975, Zadeh, 1984) is intended to represent "fuzzy" or linguistically imprecise terms, such as a "tall man", in contrast to the "crisp", well-defined events to which probability applies. For example, if T is the set of heights of "tall men", then $U_T(x)$ is a fuzzy membership function, with value 0 for x=48", 1 for x=84", and some smooth, monotonic transition from 0 to 1 for heights in between, defining the "degree of membership in T" for each x. Much more on FST can be found in the literature.

Both Myc and FST can be used to assign numbers to indicate uncertainty about propositions ("Bill is tall"), and to implication rules ("If Bill is tall and Bill is strong, then Bill is heavy"). They use rules of similar form to obtain strengths for conjunctions, disjunctions, and modus ponens implication. The strength of a proposition is attenuated by the strength of the implication rule to obtain the strength for the conclusion. This process may be repeated all the way along a rule proposition tree, to propagate uncertain beliefs from its leaves (data nodes) to its roots (conclusions).

### Dempster-Shafer Belief Functions

Dempster-Shafer Theory (DST) is designed to handle cases where the probability distribution is incompletely known. Shafer (Shafer, 1984) gives the following illustration. We've asked Fred if the streets are icy. He replies "No", and we know that 80% of the time he speaks accurately and honestly, and 20% of the time he speaks carelessly, saying whatever comes to mind. Without a prior on the proposition "the streets are icy", or a conditional probability that Fred is correct when speaking carelessly, a naive Bayesian would be unable to produce a posterior probability that the streets are icy, given Fred's answer. With $t_1$ = "the streets are not icy", $t_2$ = "the streets are icy", Shafer derives a belief of 0.8 in $\{t_1\}$, 0 in $\{t_2\}$, and 0.2 in $\{t_1,t_2\}$, meaning we have 0.8 units of evidence for $t_1$, none for $t_2$, and 0.2 for $t_1$ or $t_2$.

Shafer defines a *frame of reference* as the set of basic events and all its possible subsets, in this example T = $\{\{t_1\},\{t_2\},\{t_1,t_2\}\}$. It is convenient to use the notation $d(\tau)$ = [a,b] for a subset $\tau$ of T, where a = bel($\tau$), b = bel(T-$\tau$), i.e. a is strength of evidence for $\tau$ (its *support*) and b is strength of the evidence against it. In the example $d(\{t_2\})$ = [0. .2] meaning nothing for it and .8 against it. A Bayesian would try to allocate the remaining .2 between $t_1$ and $t_2$, but Shafer's point is that, if we don't have the information (priors etc.) we can and should leave it open. Complete lack of evidence is analogous to [0,0]; a probability of 0.5 is analogous to [0.5,0.5]. Thus DST is a generalization of probability, which allows *lack of evidence* to be represented in addition to probabilistic belief. DST also provides methods such as Dempster's rule for combining evidence from different sources (Shafer, 1976).

Unfortunately, when a frame of reference S has m propositions, and $n = 2^m$ basic events, there are $2^n$ subsets in that frame. Using such a belief function could potentially be a burden that is exponentially greater than the full probabilistic representation. However, there are arguments that approximations can be used, and an important special case where a simple exact function can be used. Unfortunately the special case is a strong argument against the approximations, as we shall argue below.

## Comparing UISs

To answer questions about differences in performance, a common interpretation of the inputs and results is required to make them commensurable. Presumably, the ultimate purpose of any expert system is to lead to better decision. If two different representations of uncertainty lead to making the same decision, then they are operationally equivalent. According to Bayesian Decision theory, decisions can reveal probabilistic beliefs about the outcomes on which the decisions are based. Even if the decision-maker doesn't actually have probabilistic beliefs, if he chooses coherently he will act as though he has them. So in principle, if a non-probabilistic approach to uncertainty provides an integrated theory of how to make decisions based on its representations of uncertainty, then this would imply an operational correspondence between uncertain beliefs expressed in the probabilistic and non-probabilistic forms, where they produced the same decision. This would allow direct comparison of the representations. Notably, however, the non-probabilistic approaches do not provide agreed upon decision strategies, and so this is unfortunately impossible.

Nevertheless, there are obvious, simple ways to make transformations from at least FST and Mycin to probability. For FST one can simply estimate probability by the fuzzy membership function:

$$p(x) = f(x)$$

This corresponds to the interpretation of f = $U_S(x)$ as "the probability that x will be classified as S, by the expert after whom the system is modelled, is f". The FST combination rules then form a quasi-probabilistic model of the expert's reasoning.

For Mycin, the conversions are taken from the verbal definitions of Certainty Factors, m(x) (Shortliffe, 1976): 1 means true, so p(x) = 1, 0 means no evidence, so we assume the prior, p(x) = p'(x), and -1 means definitely, so p(x) = 0. We use piecewise interpolation between these three points of correspondence.

DST can be looked at as providing bounds on probability. The belief in an event set $\tau$, a = Bel($\tau$) and in its complement, b = Bel(T-$\tau$), provide lower and upper bounds on its probability. We will derive and discuss some problems with this below.

### Extra parameters and assumptions about correlations

As we mentioned, the number of parameters of a joint probability distribution over m propositions, has $2^n - 1$ free parameters. In general, the number of conditional probabilities, specifying rule strengths, and marginal probabilities, specifying prior beliefs, which elicited from experts, will be far fewer than the number of parameters, and hence insufficient to completely the joint distribution.

Any uncertain inference method, by implication at least, makes certain assumptions about the unspecified parameters, particularly the correlation between propositions. For example, the combination rules for Fuzzy Set Theory and for Mycin (under some conditions on the prior) are equivalent to probabilistic inference assuming *maximum correlation* between the propositions. One can also use simplified probabilistic inference methods that assume *minimum correlation* or *independence* between the input propositions. For convenience we will label these assumptions, "MaxC", "MinC", and "Ind" respectively. The following table gives the combination rules for calculating the probability of the conjunction and disjunction of two events, A and B, as a function of their probabilities, $p(A)$ and $p(B)$, corresponding to each assumption. It also gives "probabilistic modus ponens", that is, $p(B)$, as a function of $p(A)$, and the "rule strength" or conditional probability, $p(B|A)$. In the latter case, the assumptions about correlation or independence are subject. to $p(B|A)$, and are assumed to constrain $p(B|\neg A)$. In particular MaxC (and FST) implies that $p(B|\neg A)=0$, MinC implies $p(B|\neg A)=1$, and we assume that Ind lies between them and implies $p(B|\neg A)=0.5$. To avoid these somewhat arbitrary assumptions Prospector (Duda et al, 1976) allows explicit specification of this "lower strength" for each rule $p(B|\neg A)=0$.

For maximum correlation (MaxC):

$$\begin{aligned} &p(A\&B)=Min(p(A),p(B)),\\ &p(AorB)=Max(p(A),p(B)),\\ &p(B)=p(B|A)\times p(A) \end{aligned} \quad (1)$$

For Minimim correlation (MinC):

$$\begin{aligned} &p(A\&B)=Max(0,p(A)+p(B)-1),\\ &p(AorB)=Min(1,p(A)+p(B)),\\ &p(B)=p(B|A)p(A)+1-p(A) \end{aligned} \quad (2)$$

For "Independence" (Ind):

$$\begin{aligned} &p(A\&B)=p(A)p(B),\\ &p(AorB)=p(A)+p(B)-p(A)p(B),\\ &p(B)=p(B|A)p(A)+[1-p(A)]/2 \end{aligned} \quad (3)$$

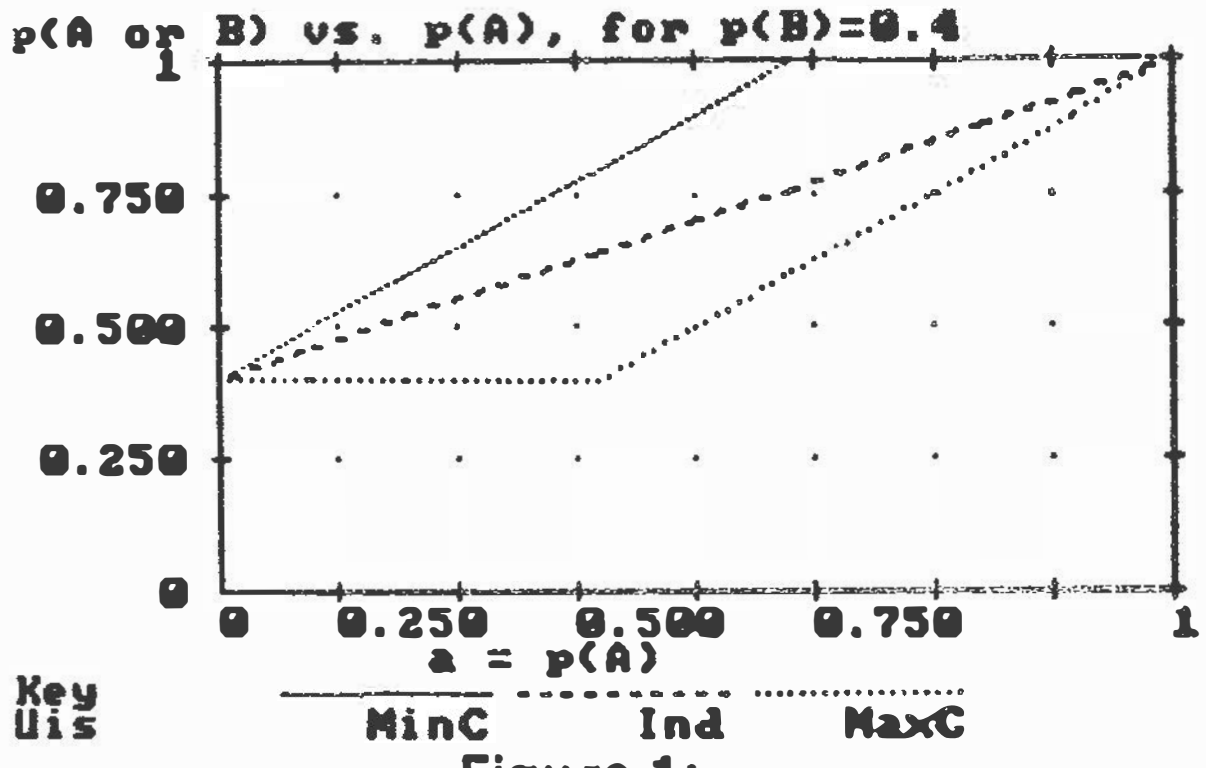

Figure 1:

To visualize the possible effect of this range of assumptions about the correlations we have graphed some implications of these rules in Figures 1 to 4. Figure 1 shows $p(AorB)$ as a function of $p(A)$ for $p(B)=.4$, and Figure 2 shows $p(A\&B)$ as a function of $p(A)$ for $p(B)=.6$. In both these cases Mycin and Fuzzy Set Theory make the identical assumption to MaxC, that is maximum correlation. These give the maximum probability for the conjunction and minimum for the disjunction. The minimum correlation assumption, "MinC", does the reverse. The range of possible error increases as A and B become more uncertain (nearer 0.5).

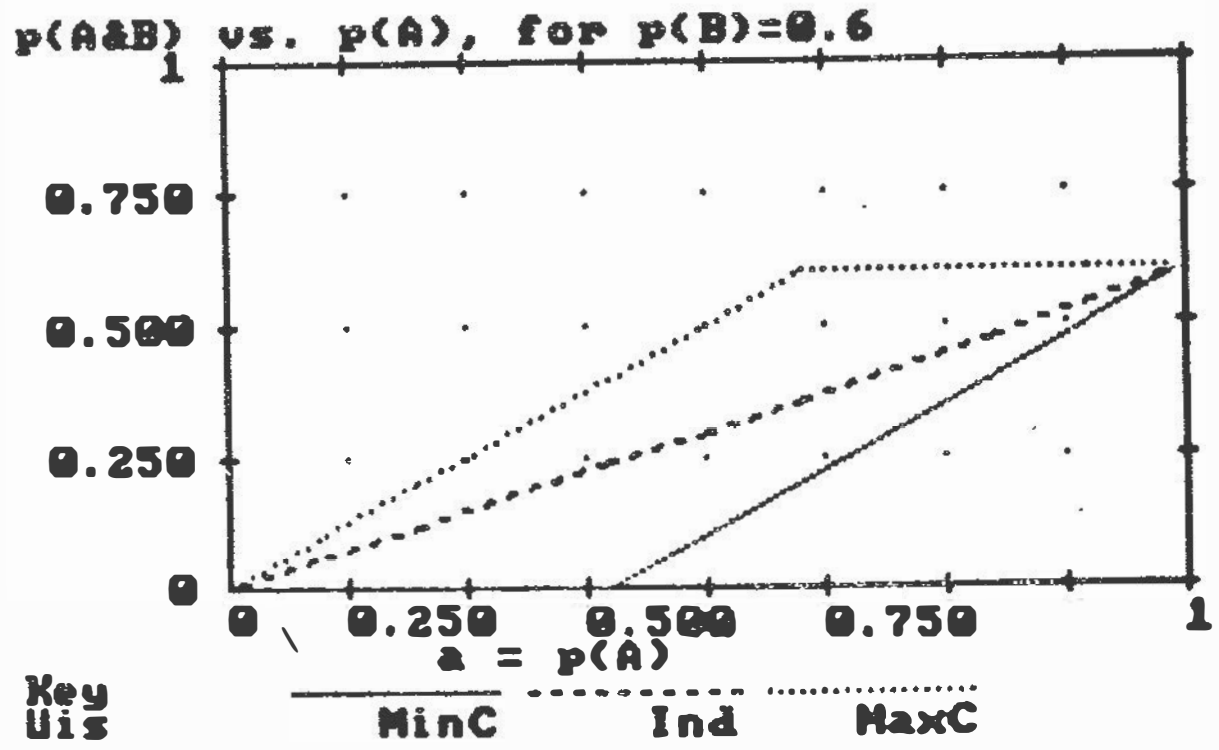

Figure 2:

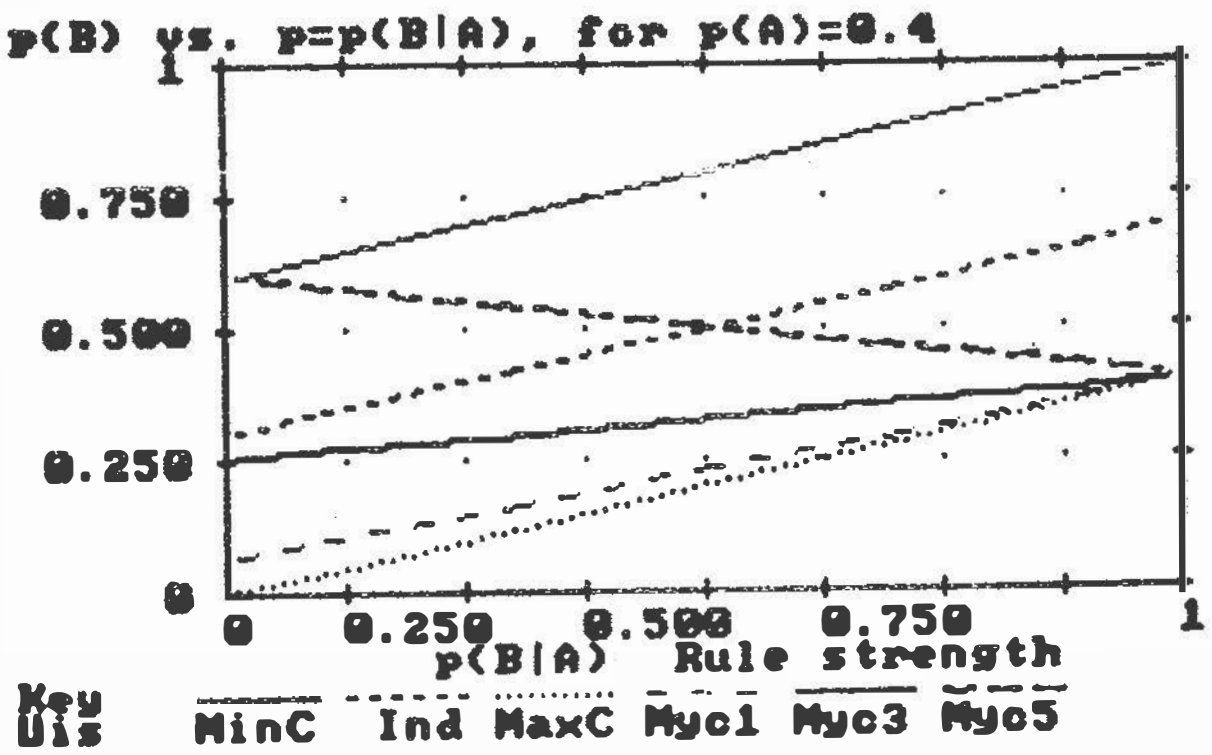

Figure 3:

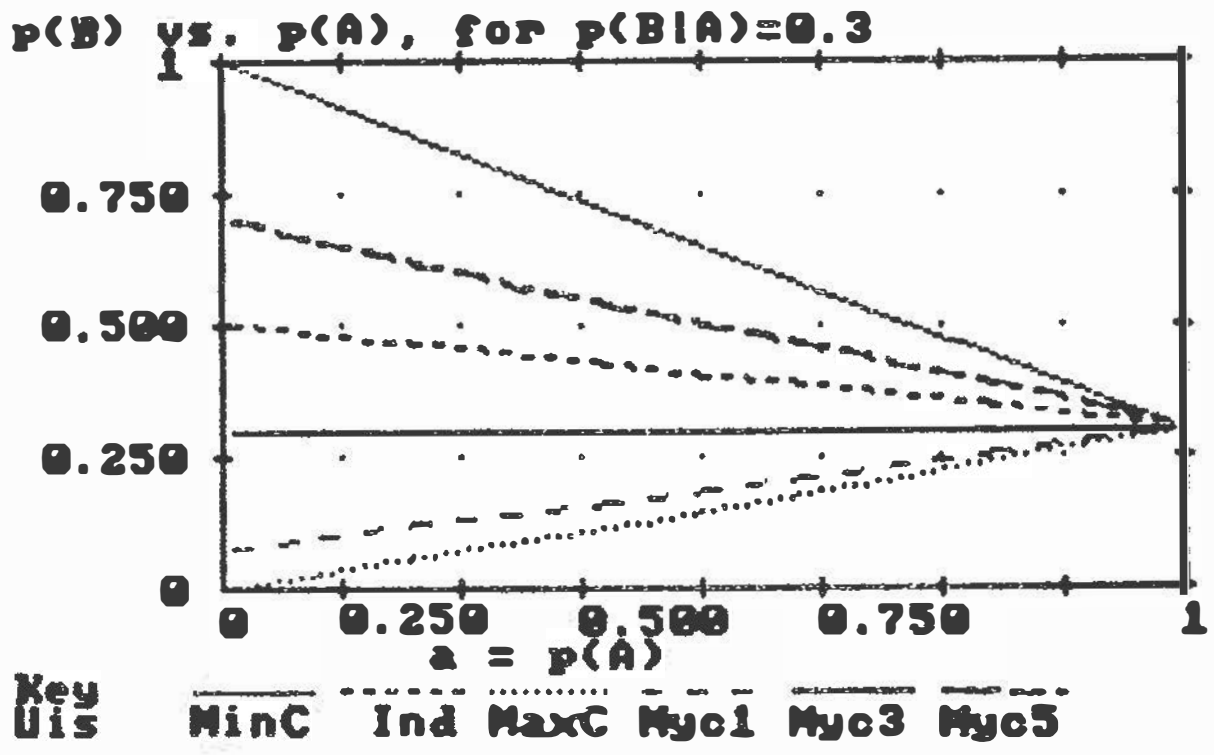

Figure 4:

Figure 3 shows the probability of conclusion C, as a function of rule strength $p(B|A)$, for the probability of the antecedent, $p(A)=0.4$, and Figure 4 shows it as a function of $p(A)$, for $p(B|A)=0.3$. In these cases Fuzzy Set Theory is still identical to MaxC, but Mycin is assumes that if the antecedent is false then p(B) is unaffected and remains at its prior $p(B|\neg A)=p_0(B)$. Since Mycin depends on what prior probabilities are used we show three examples, "Myc1", "Myc3" and "Myc.5" which assume all priors are 0.1, 0.3 and 0.5 respectively. Under all conditions, Maxc and Mycin give the lowest probability of the conclusion. The range of error becomes very large for small *p(A)*, for any rule strength. Whereas the slope of *p(B)* with respect to the rule strength is similar for the three probabilistic UIS's, MinC, MaxC and Ind, the slope with respect to *p(A)* is larger for MaxC and Mycin with small priors. Indeed Figure 4 shows that the slopes can have different sign. FST can respond in the opposite direction to the Ind assumptions, and so can Mycin if the rule strength is less than its prior. Under these circumstances alternative assumptions can lead to qualitatively quite different behavior.

### The maximum entropy principle

Any assumptions, such as those discussed above, of correlation or absence of correlation imply additional information that has not in fact been given. So in general they have no basis. One approach to this problem is to explicitly minimize the additional information implied by choosing the prior distribution that maximizes entropy subject to the constraints of the specified rule-strengths and probabilities. Essentially the maximum entropy distribution makes the weakest possible assumptions consistent with the specified knowledge. Similarly, when additional information is obtained on a specific case, the prior can be updated to incorporate it by minimizing the cross-entropy (MXE) between the prior and posterior distributions. Cross-entropy is a measure of how much information one would have to receive to change one distribution into another[1].

Shore and Johnson (Shore and Johnson, 1980) have proved that this MXE approach has several very desirable properties. It produces a unique result that is invariant under any one-to-one transformation of the co-ordinate system for specifying events, and under changes in between equivalent forms for specifying the joint distribution of pairs of independent events and subsystems. Any other method for filling in the distributions, subject to linear constraints (such as rule- and data-strengths), and which satisfies these weak and highly desirable conditions would give the same results, and so be identical in effect.

Of course, the estimate need not be computed via the general MXE calculation. Indeed, there are several important special cases: assuming independence when no correlation data is provided (a rule-strength is information on how the antecedant and consequent are correlated), ordinary conditioning when probabilities are provided for disjoint events (including Bayes' Theorem). For rule- and data-strengths, the log-linear model is formally quite similar, and has been used with good results in several projects.

In summary, the use of other UIS gives results operationally equivalent to using various assumptions within BDT, but by the above two arguments, MXE gives the only estimate consistent with BDT. Hence, using UIS which are not operationally equivalent to MXE will incur losses from violations of BDT. Using calculus of variations and simple numerical search algorithms, we can calculate the appropriate maximum entropy distributions for small problems quite simply, albeit slowly, and so are able to find the MXE conclusions as well as those of the other UIS, and so can compare them to assess how large or small the incurred losses are.

### Comparing UIS behavior for a simple rule

Probably the most typical form of rule in rule-based systems is where a consequent is dependent on the conjunction of a set of antecedent propositions. To obtain some insights into how various UIS's compare with each other and with the maximum entropy method (MEP), it is useful to explore a simple rule with two antecedents. We express the consequent, C as conditioned on the two antecedents, A and B, as $p(C|A\&B)=p$, and examine the behavior of *p(C)* as a function of *p(A)*, *p(B)* and *p*.

In this case we will compare MEP, Mycin, FST and Ind. The FST assumes maximum correlation between A and B, and A&B and C, in equations (1). Likewise Ind assumes independence between each pair. The MEP method first calculates the prior distribution that maximizes entropy of the joint distribution *p(A,B,C)*, subject to the given conditional probability, and then obtains the posterior distribution given the marginals *p(A)* and *p(B)*, with minimum cross-entropy to the prior. For a fair comparison Mycin uses the same priors as the MEP distribution, but updates them using Mycin combination rules.

---

[1]Thus, although Zadeh criticizes UIS's other than FTS for "assuming away" missing or ambiguous data, the FST algorithms he advocates are operationally equivalent to making stronger assumptions than necessary, by virtue of the fact that they give different results than the minimum cross entropy, or minimum assumption, method.

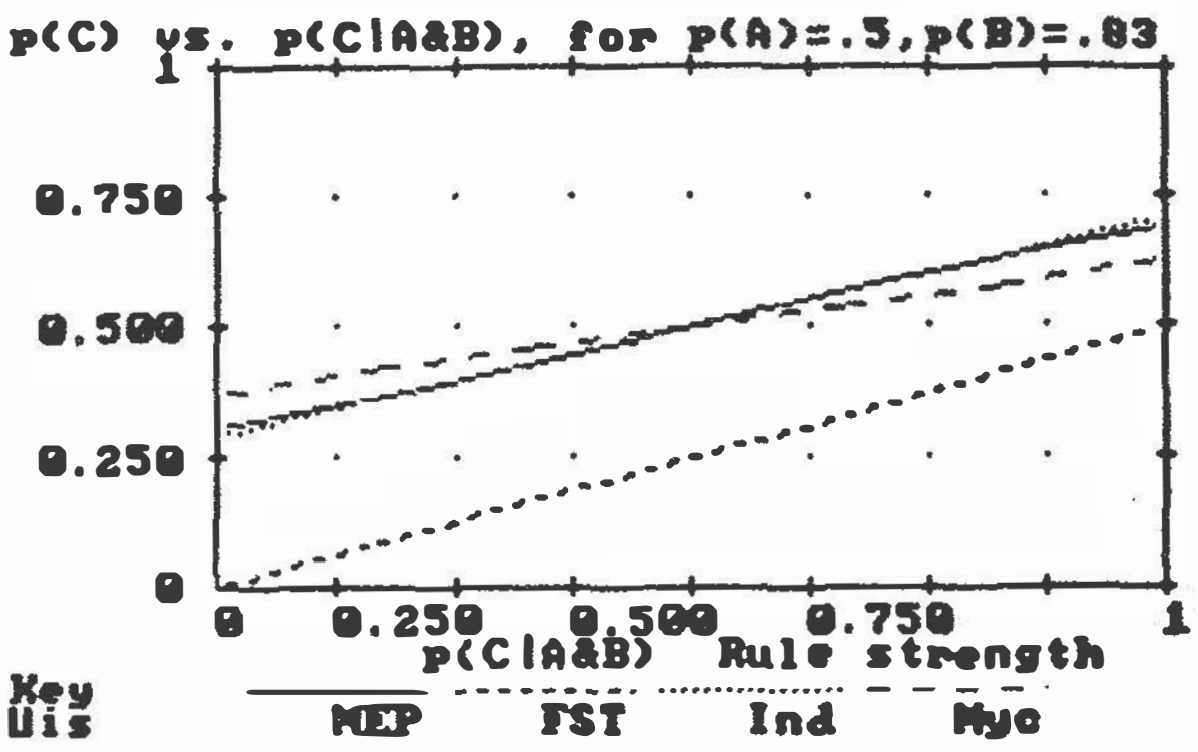

Figure 5:

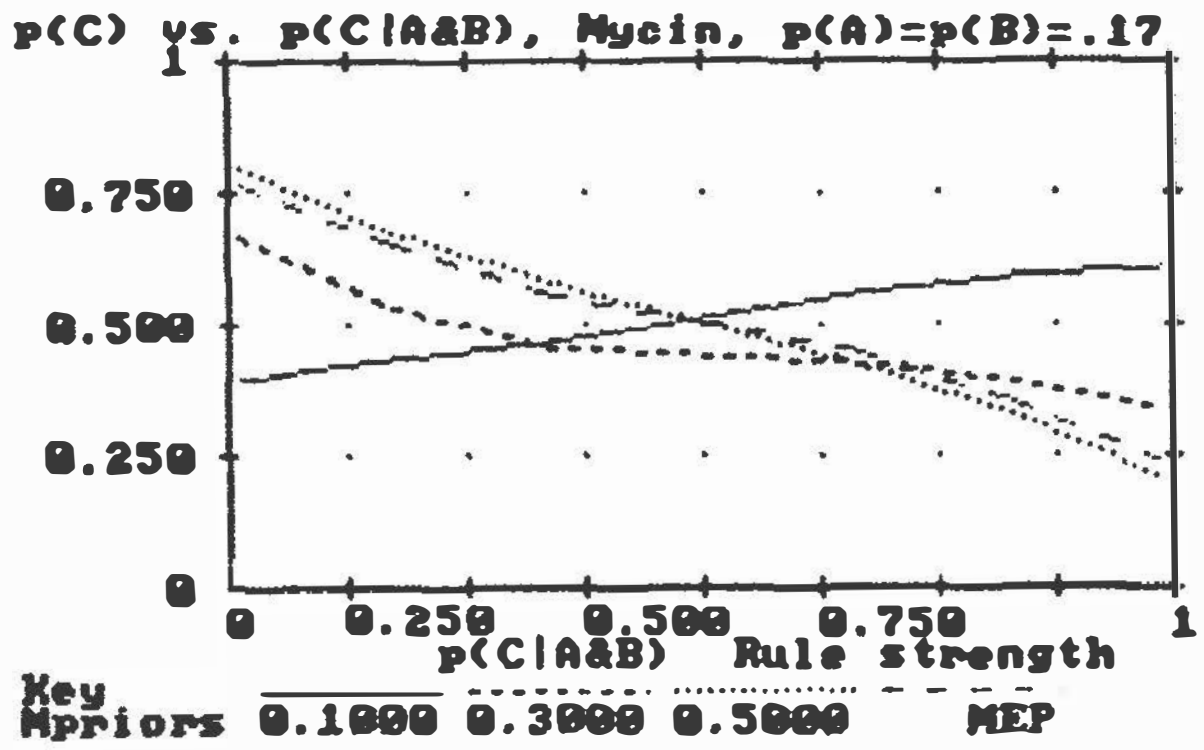

Figure 6:

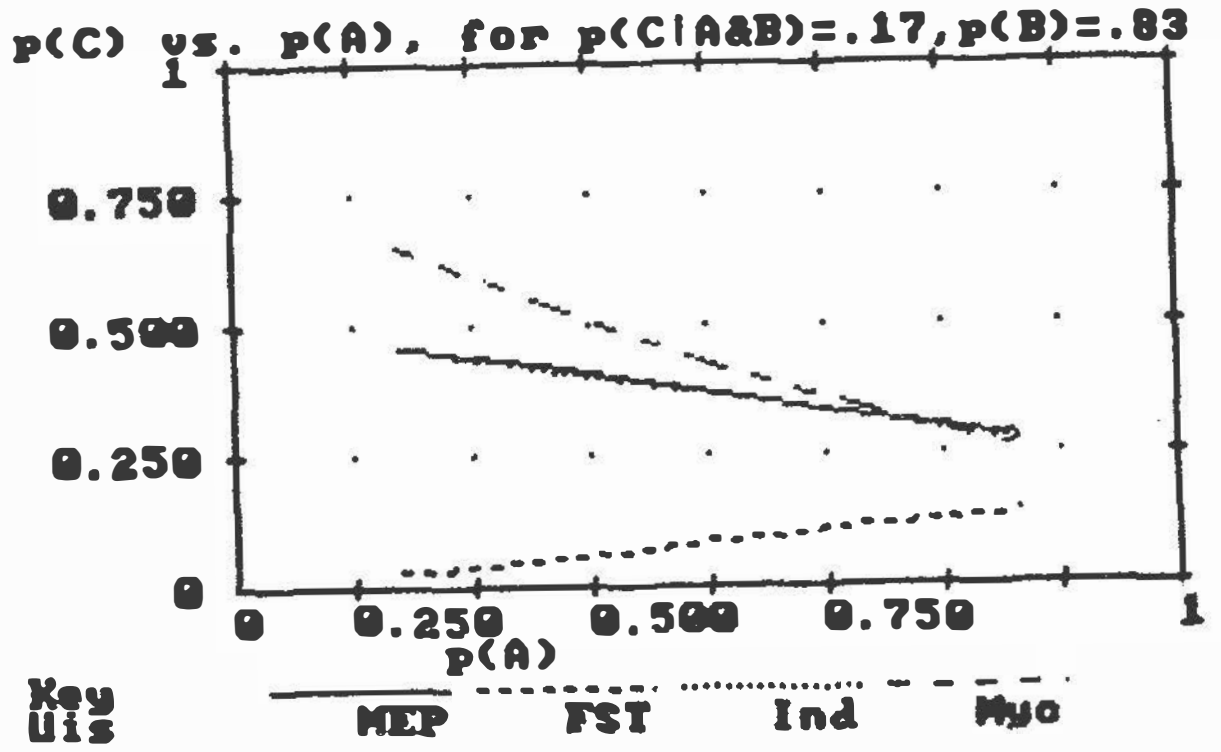

Figure 7:

Figures 5 and 6 show *p(C)* as a function of rule strength $p = p(C/A\&B)$, for fixed values of *p(A)* and *p(B)*. Figure 7 shows *p(C)* as a function of *p(A)* for fixed *p* and *p(B)*. (A and B enter symmetrically and so *p(A)* and *p(B)* could be interchanged.) In all cases Ind is very similar to MEP (and so is largely obscured by it in the graphs). The methods are most similar for large *p(A)* and *p(B)*, having similar slope with respect to *p* for $p(A)>.5$, $p(B)>.5$ as in Figure 5. In all three cases FST gives the lowest probability *p(C)*, but the largest slope with respect to *p* and to *p(A)*. Note that FST gives a positive slope with respect to *p(A)* for $p<0.5$, and $p(A)<p(B)$, as in figure 7, even when all other methods give *p(A)* negative slope. As illustrated in Figure 6, Mycin, unlike the others, can give a negative slope with respect to p, implying that the stronger the rule the less likely the consequent. This happens for all $p(A)<p_0(A)$. Note that if the priors are all very small, as may be true in some applications, then the Mycin certainty factors and rule strengths will remain positive, and hence this counter-intuitive behavior will be avoided. But in general both FST and Mycin can behave in qualitatively different fashion to the other methods.

## A problem with Dempster-Shafer Theory

The definition of Dempster-Shafer Belief functions suggests an obvious correspondence with probability, with the Belief functions providing upper and lower limits on the probability. But Shafer resists this interpretation, and in fact it turns out that the results of making this correspondence are trivial. This arises from an unexpected consequence of the definition of the belief function. To explain this we must first give a more formal definition.

In general, Shafer utilizes four objects to derive a belief function: Two frames of reference, S and T, each of which contains a set of basic events and all its possible subsets, a probability distribution over the basic events in S, $p_S$, and a compatibility relation C such that sCt iff s can be true in S simultaneous with t being true in T. For $s \in \sigma \subseteq S$ and $t \in \tau \subseteq T$, Shafer defines a belief in $\tau$ via (4)

$$Bel(\tau) = p_S\{s|(\forall t)(sCt \rightarrow t \in \tau)\} \qquad (4)$$

Let us now consider an interesting special case of DST, by returning to Shafer's "icy" example. Let $s_1$ = "Fred was careful and honest", $s_2$ = "Fred was careless", $t_1$ = "streets are not icy", $t_2$ = "streets are icy"; we know $p(s_1) = 0.8$, $p(s_2) = 0.2$, and $s_1$ and $t_2$ are incompatible. The compatability relation between S and T is shown in figure 8, along with $p_{ST}$. We know that $p_{ST}$ must have the form given, but without more data a naive Bayesian could not calculate a precise value for $\alpha$. Let us consider, however, that there is a remote possibility of $(s_1\&t_2)$ happening - Fred misheard the question, we misheard the answer, it has since cooled, or something else. This gives figure 9

The new Bel function gives DST($\{t_1\}$) = [0,0], DST($\{t_2\}$) = [0,0], DST($\{t_1,t_2\}$) = [1,0], no matter how small $\beta$ is. This discontinuous h sensitivity to $p_{ST}$ suggests that ignoring "negligable events" will generally have radical effects.

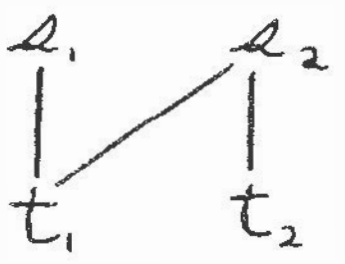

| | $s_1$ | $s_2$ |
|---|---|---|
| $t_1$ | .8 | .2α |
| $t_2$ | 0 | .2(1-α) |

Figure 8: Shafer's example

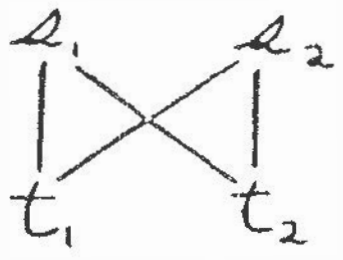

| | $s_1$ | $s_2$ |
|---|---|---|
| $t_1$ | .8-β | .2α |
| $t_2$ | β | .2(1-α) |

Figure 9: Shafer's example modified

The pathology arises in the following general case. Equation (4) implies equation (5).

$$Bel(\tau) = p_S\{s|(\forall t)(p_{ST}(t|s)>0 \rightarrow t \in \tau)\} \quad (5)$$

$$Bel(\tau) = max\ p_S(\sigma), \quad s.t.(\forall s \in \sigma)(p_{ST}(\tau|s)=1)$$

Suppose that $p_{ST}$ does not definitively rule out any combination of $s_i$ and $t_j$, i.e. $p_{ST}: SxT \rightarrow (0,1)$. Then the only $\tau$ which is guaranteed, for any $\sigma \subseteq S$, is T itself. Hence, DST($\tau$) = [0,0] for all $\tau \subset$ T, and DST(T) = [1,0]. In BDT, $p_S(\sigma)$ is clearly a tight lower bound on $p_T(\tau)$, which can, but usually will not, be acheived. In everyday terms, this extreme conservativism requires us to completely ignore whatever anyone says, unless we literally believe them to be infallible. In the military application of processing sensor data, all data would be ignored, because individually each sensor is fallible, and there is even a remote chance that all will fail simultaneously. But ignoring reports of incoming missiles is very likely to be fatal.

Clearly, this is ridiculous, and any expert who treated improbable events equally with the probable ones would perform quite poorly. *As a Bayesian*, one can show that the expected cost of ignoring possible events declines smoothly toward zero as their probability declines toward zero. Indeed, knowing $p(t_2|s_1)<10^{-50}$ in figure 9 would surely be enough to justify ignoring the possibility of ($t_2$&$s_1$), and there is no need to seek a precise number. But to avoid using such estimates about correlations between the S and T frames was Shafer's aim, and so the Bayesians' way of handling Bel's discontinuous dependence on $p_{ST}$ is inadmissable. Of course, if we do use Bayesian reasoning to justify ignoring improbable events and weak correlations, then the problem becomes one of bounding- or sensitivity-analysis within BDT, *not* one of DST versus BDT.

## Evaluating differences in results

Given some method for tranforming the conclusions of non-probabilistic UIS's to be commensurable with probabilities, we need some way to evaluate the importance of the differences between results. If we had an explicit utility function for decisions based on the output of the inference system, then we could simply calculate the expected loss in utility for using suboptimal inference methods. But in practice explicit utility functions are often hard to come by. So we propose to use both mean absolute error and mean squared error, on the grounds of simplicity and also because of their consistency with plausible classes of utility function.

### Normalizing error estimates

A difficulty arises in comparing performance on different cases in that they are likely allow different ranges of error. For example, if we randomly guess at the probability of 0.5, it is impossible to be off by more than 0.5, but if we guess the probability of an almost certain event (probability 0 or 1), then it is possible to be off by almost 1. An error near .5 is almost the worst possible in the former case, but is about average for the second. So it is useful define normalized performance measures, which compare estimates to random guesses, and the worst possible results. The worst possible estimate of p is *max*(p,1-p). We can define the random guess value as the expected error, if the estimate were uniformly distributed over the estimator's domain. For example, for FST, the domain is the closed interval [0,1]; for MYC, the closed interval [-1,+1]; for DST, the triangular region $\{(a,b)|0 \leq a \leq b \leq 1\}$, where a is the lower bound on p and b is the upper bound. The expected mean absolute error, $\mu(|\varepsilon|)$, and mean squared error, $\mu(\varepsilon^2)$, for such random guesses are given below for different UIS's:

**FST:**

$$\mu(|\varepsilon|) = \frac{1}{2} - p(1-p) \quad (6)$$

$$\mu(\varepsilon^2) = \frac{1}{3} - p(1-p) \quad (7)$$

**MYC:**

$$\mu(|\varepsilon|) = \quad (8)$$

$$(2(p^2+p_0^2)+Q-4pp_0)/(4Q),$$

$$where\,Q=\begin{cases} p_0 & if\; p<p_0 \\ 1-p_0 & otherwise \end{cases}$$

$$\mu(\varepsilon^2)= \qquad (9)$$

$$(2p_0^2-6pp_0+6p^2+1+p_0-3p)/6$$

**DST:**

$$\mu(|\varepsilon|)= \qquad (10)$$

$$\tfrac{31}{18}(p^3+(1-p)^3)+\tfrac{7}{2}p(1-p)$$

$$-\tfrac{2}{3}(A+B)-\tfrac{11}{9},$$

$$where\begin{cases} A=p^3 ln(p) \\ B=(1-p)^3 ln(1-p) \end{cases}$$

$$\mu(\varepsilon^2)=11/36-p(1-p) \qquad (11)$$

The normalized measures rescale the errors to give 1 for zero error, 0 when as good as random guessing, and -1 for the worst possible, with linear interpolation in between. The normalized absolute error, is given by piecewise linear interpolation between:

$$\eta = +1 \; iff \; \varepsilon = 0. \qquad (12)$$
$$\eta = 0 \; iff \; \varepsilon = \mu(|\varepsilon|)$$
$$\eta = -1 \; iff \; \varepsilon = max(p, 1-p)$$

The normalized squared error, $\zeta$ is defined similarly with mid-point $\mu(\varepsilon^2)$, and worst value $max(p^2,(1-p)^2)$.

## Comparing performance on an example rule-set

All considerations given above come together in the framework diagrammed in figure 10. The experiment begins with a collection of rules, R, stated as conditional probabilities of consequents given antecedants, and prior probabilities. In this example, the rules allow inferences about the probabilities of four different kinds of nuclear reactor accidents. Aside from the conditional probabilities of accidents given symptoms, it seems plausible to include a rule that the probability of *no* accident is quite small, since the system would not be used unless alarms had sounded - but false alarms are possible. These rules are taken to be a general description of the domain, as opposed to the case-specific data, D, which describes which particular alarms and symptoms a particular accident caused. The MEP prior, $p_0$, is estimated by maximizing entropy subject to the rules R. This prior is then updated using minimum cross-entropy, by D to yield the posterior $p_1$, from which the probabilities of various accident types (i.e. the conclusions) may be read. R and D are also converted to whatever strengths the current UIS uses, giving R' and D', which might be a collection of Mycin rule-strengths and CF's. The UIS is then used to propagate the D' using the rules, R', to obtain the strengths for the conclusions, C', which are then converted back to probabilities, $p'_1$, and compared to $p_1$.

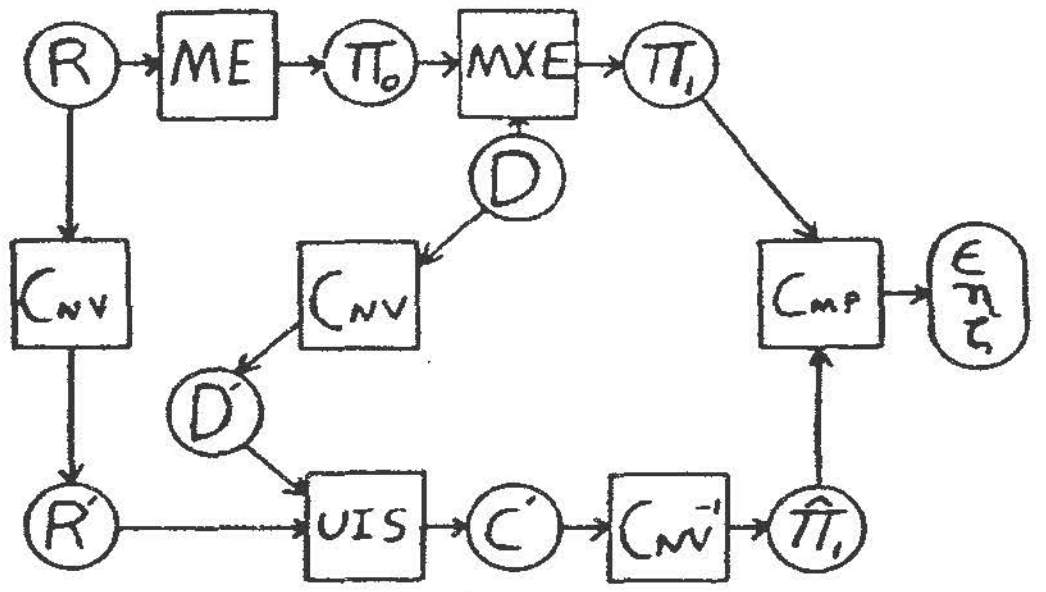

**Figure 10:** Basic Experiment Design

It is important to note that $p_1$ is estimated in two stages, by using R with maximum entropy and using D and $p_0$ with minimum cross entropy, rather than using RUD with maximum entropy. To see why, consider a case where one rule is *"given swollen belly and sickness in morning, there is a 40% chance of pregnancy"*, and another is *"given male-ness, there is a 0% chance of pregnancy"*. The data is that the patient has a swollen belly, is sick in the morning, and is a male. Whatever our prior $p_0$ is, if it is consistent with the two rules, then when it is updated with the data, it will give a probability of zero to pregnancy. Hence, *in the posterior* $p_1$, the probability of pregnancy given male-ness is zero, not 40%[2]. It would be impossible to have the rules and the data both hold in one distribution, and if possible would lead to the wrong conclusion of a 40% chance that a male was pregnant.

The original rule incorporated the assumption that there was a significant chance of the patient being female, which is directly contradicted by the new data. If rules were conditioned on not just one or two propositions, but on every possible proposition, then they would include no assumptions about prior probabilities, and it would make no difference whether $p_1$ was found in one step or two, or even conditioned several times by different D's along the way. But eliciting rules in this form may be a good deal harder.

[2] A simple example of the non-monotonic reasoning inherent in BDT.

The following nuclear reactor accident diagnosis example is presented less as a definitive analysis than an illustration of the general methods and a source of suggestions for future research. The rules are taken from (Kunz, 1984); there are 9 rules and 18 propositions, including 10 leaves (data D), 4 roots (conclusions C), and 4 intermediate propostions, I. The four conclusions are designated *A*, *B*, *C*, and *D*., as in figure 11.

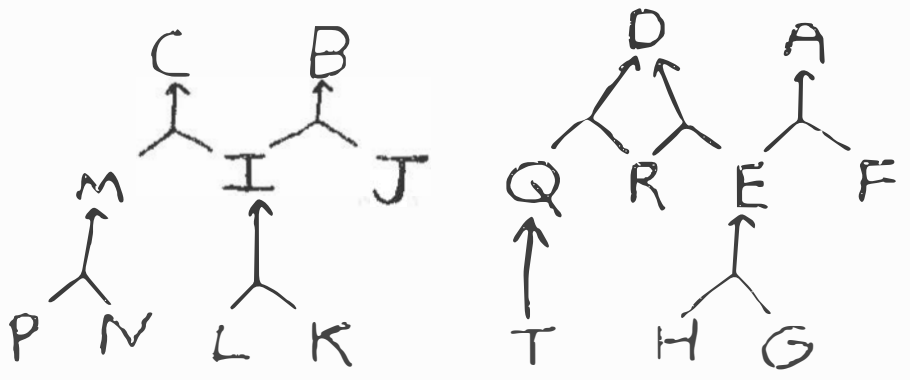

**Figure 11:** Example rule-set

We specified rule-strengths, varying from 70% to 95%. and added two new rules - that the probability of no accident was 1%, and the probability of a single accident was 95%. Hence, there was a 4% of a multiple failure. It should be noted that these two rules induce certain other correlations, which might be expressed as a cyclic set of conditional probabilities, such as $p(B|A)<0.04$.

This set of 11 rules was used to form prior $p_0$ as in figure 10, which was then used with different D's in four different tests. For each of the four accident types to be concluded, we picked out those leaves which directly or indirectly supported the conclusion and set their probabilities to 95%; all other leaves were set to 5%. For each case, the MXP posterior $p_1$ was computed for comparison with the results of the other UIS's. FST, Mycin, Ind, and DST. Averages of the absolute error $|\varepsilon|$, squared error, $\varepsilon^2$ and the corresponding normalized errors, $\eta$ and $\zeta$ were calculated for the intermediate I and conclusion C propositions, and both types together. These averages for FST, MYC, Ind and DST for one of the tests are shown in figure 12; results for the other three tests were qualitatively similar. FST tended to guess probabilities as too extreme, i.e. it treated data as supporting or disconfirming conclusions too strongly. MYC did the same, but not as much. One would expect MYC to do better than FST, even though it has similar combination rules, because it uses data from the prior which FST does not. Also, it only reasons about changes from the prior to the posterior, rather than directly about the posterior. DST always did slightly worse than chance, because its error function is influenced both by the width and the off-centeredness of the estimate, and it always had the greatest possible width. DST showed the pathology discussed earlier, and its poor results should be viewed in that light.

| | | FST | MYC | DST | IND | PRS |
|---|---|---|---|---|---|---|
| $\|\varepsilon\|$ | I | .357 | .237 | .255 | .077 | .104 |
| | C | .224 | .200 | .294 | .344 | .136 |
| $\eta$ | I | -.368 | .092 | -.144 | .702 | .592 |
| | C | .232 | .229 | -.082 | -.121 | .519 |
| $\varepsilon^2$ | I | .141 | .065 | .088 | .011 | .015 |
| | C | .052 | .044 | .127 | .122 | .023 |
| $\zeta$ | I | -.197 | .355 | -.115 | .884 | .831 |
| | C | .578 | .554 | -.072 | .106 | .786 |

**Figure 12:** Summarized comparison results

MYC is ranked better than FST for mean absolute error, but vice versa for the normalized errors. This might be interpreted as FST making better use of what information it has, but MYC slightly more than compensates by using more data. Interestingly, by all four measures the performance on C was better than on I, for FST and MYC even though one would expect errors to compound and performance to deteriorate as we propagate strengths farther.

## Final remarks

Detailed analysis of the behavior of a single rule in isolation, as outlined in the first part of the paper, can give some insights into the differences in behavior between UIS's and the conditions under which they differ, qualitatively and quantitatively. However the ultimate impact of these differences within a system of many rules depends on aggregate characteristics of the system. To understand these we must explore wider clases of rule sets, both experimentally and analytically to understand the results.

The experimental observations presented, being based on only four different tests using one rule-set, are primarily illustrative. There are obviously many interesting lines of research open at this point - using a wider variety of rule-sets, other UIS's, using other methods of correspondence between probabilities and non-probabilistic UIS's, including perhaps a fairer test of DST, and so on. The main purpose of this paper is to present and and try to justify a framework for testing the accuracy of UIS's results, ignoring for the moment issues of computational effort, clarity, or simplicity.

The debate among the partisans of different UIS's seems to be increasing in intensity, and at times verges on the acrimonious. The divergence both in philosophical underpinnings and in programmatic goals of researchers may have contributed a degree of mutual incomprehension between the various paradigms. While the difficulties in resolving such conflicts should not be underestimated (Kuhn, 1962), we believe that clearer presentation of these fundamentals and examination of the methods against the full range of criteria, including the theoretical, pragmatic issues, as well as the experimental comparison of performance explored here, could shed some needed light. Different people will have different weightings for these criteria, reflecting their different goals, and so it may never be appropriate to attempt definitive evaluation of the techniques. But in any case, better analytic and experimental evidence which compares the performance of UIS's in terms of their results, could help to provide system designers a more solid basis for choosing among them.

## Acknowledgments

This work has been supported in part by the Intelligent Systems Laboratory of the Robotics Institute, Carnegie-Mellon University, and the National Science Foundation under grants IST-8316890 and PRA-8413097.